\documentclass[runningheads]{llncs}

\usepackage[]{}

\usepackage{eccvabbrv}

\usepackage{graphicx}
\usepackage{booktabs}

\usepackage[accsupp]{axessibility}

\usepackage[pagebackref,breaklinks,colorlinks]{hyperref}

\usepackage{amssymb}
\usepackage{float}
\usepackage{multirow}
\usepackage[misc]{ifsym}

\begin{document}
\title{Intelligent Artistic Typography: A Comprehensive Review of Artistic Text Design and Generation} 

\titlerunning{Intelligent Artistic Typography}

\author{Bai Yuhang{} \and Huang Zichuan{} \and Gao Wenshuo \and Yang Shuai{} \and Liu Jiaying}

 \institute{Wangxuan Institute of Computer Technology, Peking University
}

\maketitle

\begin{abstract}
  Artistic text generation aims to amplify the aesthetic qualities of text while maintaining readability. It can make the text more attractive and better convey its expression, thus enjoying a wide range of application scenarios such as social media display, consumer electronics, fashion, and graphic design. 
Artistic text generation includes artistic text stylization and semantic typography. Artistic text stylization concentrates on the text effect overlaid upon the text, such as shadows, outlines, colors, glows, and textures. 
By comparison, semantic typography focuses on the deformation of the characters to strengthen their visual representation by mimicking the semantic understanding within the text. 
This overview paper provides an introduction to both artistic text stylization and semantic typography, including the taxonomy, the key ideas of representative methods, and the applications in static and dynamic artistic text generation.
Furthermore, the dataset and evaluation metrics are introduced, and the future directions of artistic text generation are discussed. A comprehensive list of artistic text generation models studied in this review is available at \url{https://github.com/williamyang1991/Awesome-Artistic-Typography/}
.
  \keywords{Artistic text \and text effect \and semantic typography \and kinetic typography \and style transfer \and AIGC}
  
\end{abstract}

\section{Introduction}
\label{s-section1}

Artistic text generation focuses on turning text into visual forms that increase their artistic expression or convey their meaning. It can integrate plain text with fantastic style, decoration, and appearance, creating typography that is legible, readable, and appealing.
Such integration of visual representation and semantic understanding, not only attracts viewers, but also emphasizes the messages' meaning and strengthens the impact, making artistic text generation prevalent in graphic design, manga and comic book industry, advertisement, websites, consumer electronics, and social media.

\begin{figure}
    \centering
    \includegraphics[width=1\linewidth]{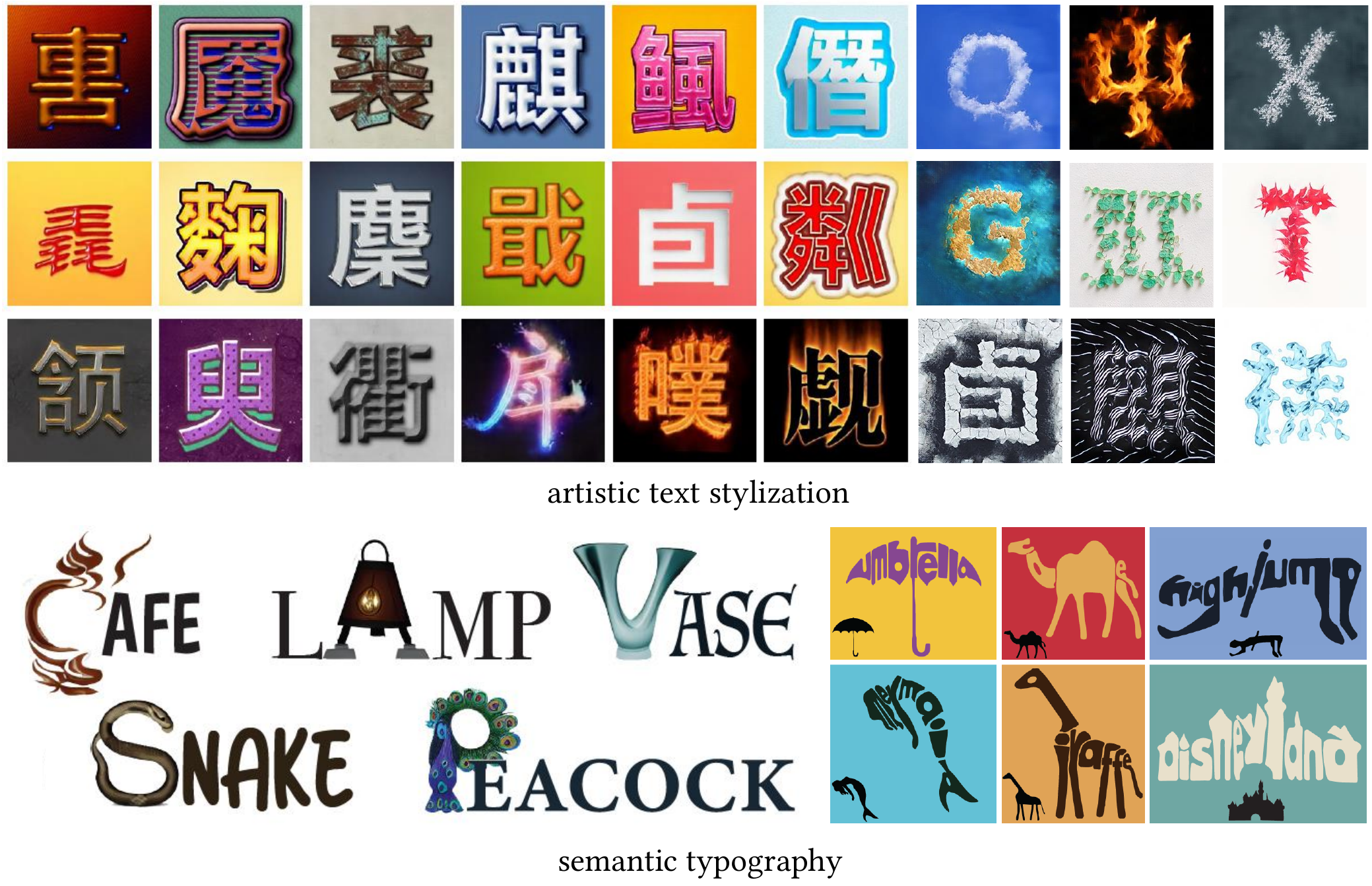}
    \caption{Artistic text generated by TET-GAN~\cite{yang2019tet}, ShapeMatching GAN~\cite{yang2019shapematching}, DS-Fusion~\cite{tanveer2023ds} and Zou \etal~\cite{zou2016legible}. Artistic text generation can be broadly classified into two categories: artistic text stylization and semantic typography. }
    \label{fig:teaser}
\end{figure}

Artistic text generation can be broadly classified into two categories: 
artistic text stylization and semantic typography. 
The former primarily involves applying visual effects (\ie, text effects) to text, while the latter focuses on redesigning the shape of text to match specific objects, as illustrated in Figure~\ref{fig:teaser}. 
Furthermore, in the era of mobile internet and multimedia, incorporating motion into artistic text to create dynamic artistic text has gained significant attention due to its captivating nature. 
However, the manual creation of such typographical art poses considerable challenges: it demands substantial time and effort. Consequently, with the advancement of computer technology, computer-assisted and even fully automated approaches for artistic text rendering and design have emerged.

Artistic text rendering pertains to artistic image rendering, a powerful tool for the automated generation of artistic images. 
The field of artistic image rendering has a long research history, starting from early methods based on traditional stroke-based~\cite{litwinowicz1997processing,hertzmann1998painterly,hertzmann2001paint} and patch-based~\cite{jacobs2001image,efros2001image,frigo2016split} approaches, to the era of deep learning with techniques like Neural Style Transfer~\cite{gatys2016image,johnson2016perceptual} and Generative Adversarial Networks (GANs)~\cite{goodfellow2020generative,isola2017image,zhu2017unpaired}. In the current era of AI-Generated Content (AIGC), empowered by the powerful large-scale models~\cite{ho2020denoising,rombach2022high}, we are now capable of generating highly fascinating artistic images. 
However, the text differs significantly from natural images or real artwork. 
First, text is highly abstract and lacks inherent visual semantic information. 
Second, the text needs to maintain legibility. 
Last, artistic text requires exquisitely designed interaction and layout to harmonize with surrounding text and background visual elements.
Designing artistic text poses unique challenges that require specialized attention.

While there have been some works focusing on artistic text rendering, a comprehensive review of these techniques and analyses of their strategies to overcome the aforementioned challenges are still lacking. This paper aims to fill this gap by providing a comprehensive overview and analysis of the current state-of-the-art techniques in the field of artistic text rendering. We aim to provide researchers with a clear understanding of the development trajectory of this topic, its potential applications, available datasets, evaluation metrics, and future research directions.

The rest of the paper is organized as below.
We begin by defining and classifying the task of artistic text generation in Section~\ref{sec2}. 
Next, Section~\ref{sec3} and Section~\ref{sec4} introduce the representative methods for artistic text stylization and semantic typography, respectively, providing detailed explanations of their main ideas and discussing their strengths and weaknesses. 
Then, Section~\ref{sec5} discusses several application scenarios for artistic text and Section~\ref{sec6} introduces the existing artistic text datasets and evaluation metrics. Finally, we discuss future research directions in Section~\ref{sec7}, and concluding remarks are given in Section~\ref{sec8}.

\section{Task Formulation}
\label{sec2}

Artistic text generation is a conditional image generation problem.
It typically involves a text input $T$ and a style input $S'$, intending to generate the corresponding artistic text $T'$, preserving the shape of $T$ while incorporating the style from $S'$.

\begin{figure}
    \centering
    \includegraphics[width=0.8\linewidth]{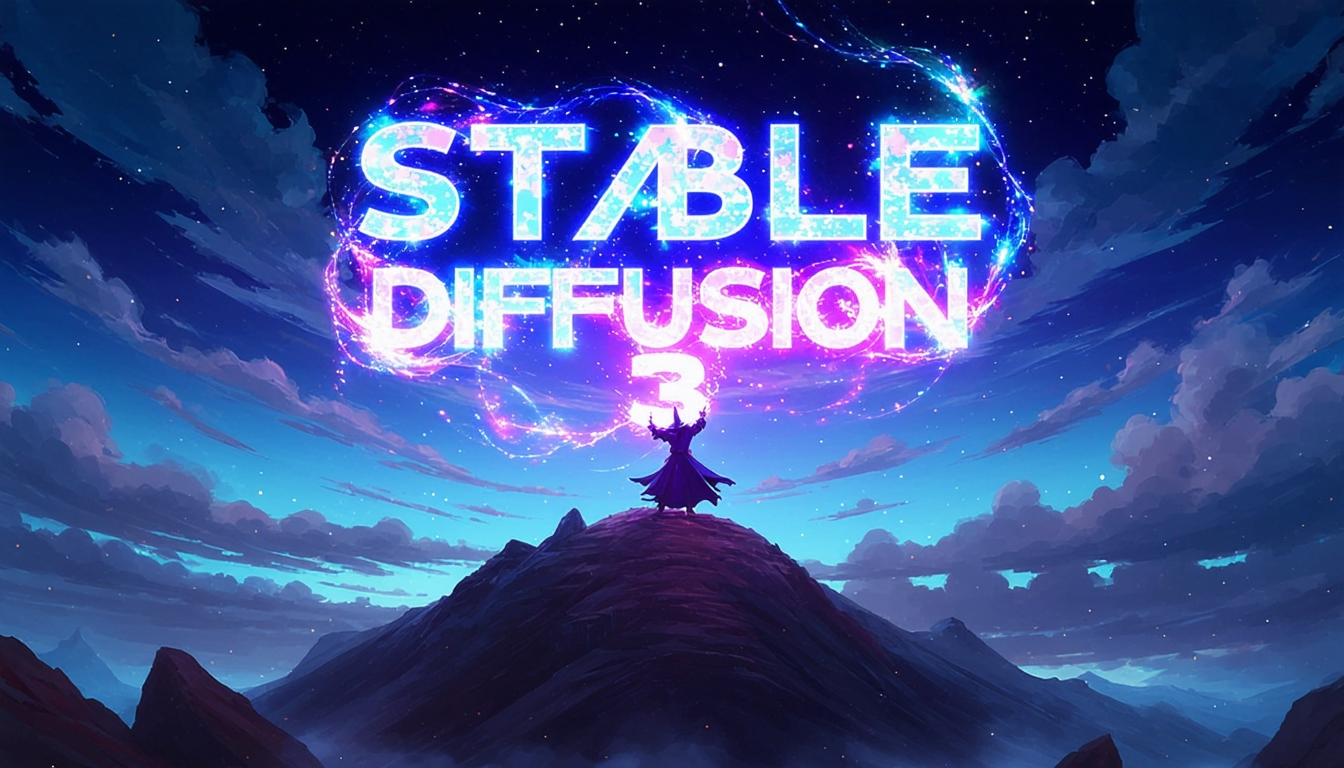}\vspace{-2mm}
    \caption{Artistic text generated by Stable Diffusion 3 with the prompt: Epic anime artwork of a wizard atop a mountain at night casting a cosmic spell into the dark sky that says ``Stable Diffusion 3'' made out of colorful energy. Image credits: Stable Diffusion 3 (\url{https://stability.ai/news/stable-diffusion-3})\vspace{-4mm}}
    \label{fig:sd3}
\end{figure}

In terms of the text input, $T$ usually can be a text raster or vector image.
With the recent development of cross-modality diffusion models~\cite{rombach2022high}, prompts can also be used as $T$ to specify the desired text to be generated.
For example, Stable Diffusion 3 demonstrates its powerful generative ability with an image generated using the prompt ``Epic anime artwork of a wizard atop a mountain at night casting a cosmic spell into the dark sky that says `Stable Diffusion 3' made out of colorful energy'' as shown in Figure~\ref{fig:sd3}.

\begin{figure}
    \centering
    \includegraphics[width=1\linewidth]{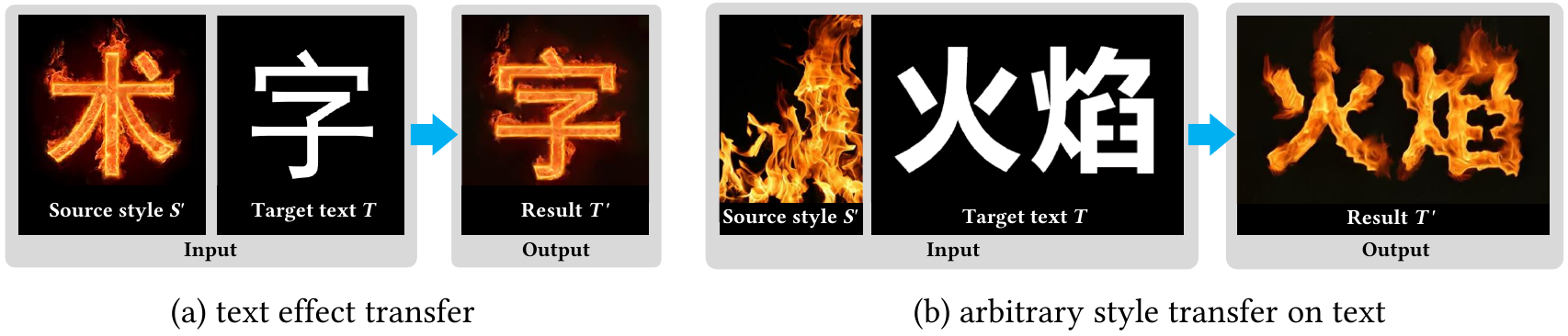}\vspace{-2mm}
    \caption{Artistic text stylization can be divided into (a) text effect transfer and (b) arbitrary style transfer on text based on whether the style input $S'$ is a well-designed text effect image or an arbitrary style image. Image credits: T-Effect~\cite{yang2017awesome} and UT-Effect~\cite{yang2018context}.\vspace{-3mm}}
    \label{fig:text_effect_class}
\end{figure}

In terms of the style input, artistic text primarily encompasses two kinds of stylish effects: one is overlaid upon the text and the other is applied to the shape of the text itself. 
Therefore, artistic text generation can be divided into two major categories: artistic text stylization and semantic typography:
\begin{itemize}
    \item \textbf{Artistic Text Stylization}. Artistic text stylization focuses on migrating visual effects (\ie, text effects) from $S'$ to $T$, including basic effects such as color, shadows, outlines, and gradients, as well as complex texture effects like flames and water ripples. Some examples are shown in Figure~\ref{fig:teaser}. 
    Here, $S'$ is usually a style image. Based on $S'$, artistic text stylization can be further divided into two subcategories: text effect transfer, which directly imitates pre-designed text effects $S'$ by artists, and arbitrary style transfer on text, which utilizes style elements from any image $S'$ to design entirely new text effects and is more similar to standard image style transfer tasks, as illustrated in Figure~\ref{fig:text_effect_class}.
    \item \textbf{Semantic Typography}. Semantic typography primarily deals with the shape of the text, aiming to deform the text into a target semantic content. For example, transforming the letter `S' in the word ``SNAKE'' into the shape of a snake or transforming the entire word ``umbrella'' into the shape of an umbrella, as illustrated in Figure~\ref{fig:teaser}. Such shape transformations enable the text to visually match its intended meaning, enhancing its expressiveness and making the conveyed message more accessible even to those unfamiliar with the language. 
\end{itemize}
It is worth noting that these two major approaches are not mutually exclusive. They can be combined to simultaneously modify the shape and apply visual effects to create more fascinating artistic text artwork.

Meanwhile, according to the modality, the task can be divided into static artistic and dynamic text generation. Static artistic text generation primarily focuses on generating still images of artistic text. It involves applying various static text effects and style elements to the text, commonly used for creating visually appealing typographic designs, logos, posters, and other static visual compositions. 
In the context of the multimedia era, incorporating motion into artistic text has gained significant attention. Dynamic artistic text generation involves generating videos or GIF animations that showcase animated artistic text. 
In this case, there are two main aspects to consider: text effect animation and text shape animation.
\begin{figure}
    \centering
    \includegraphics[width=1\linewidth]{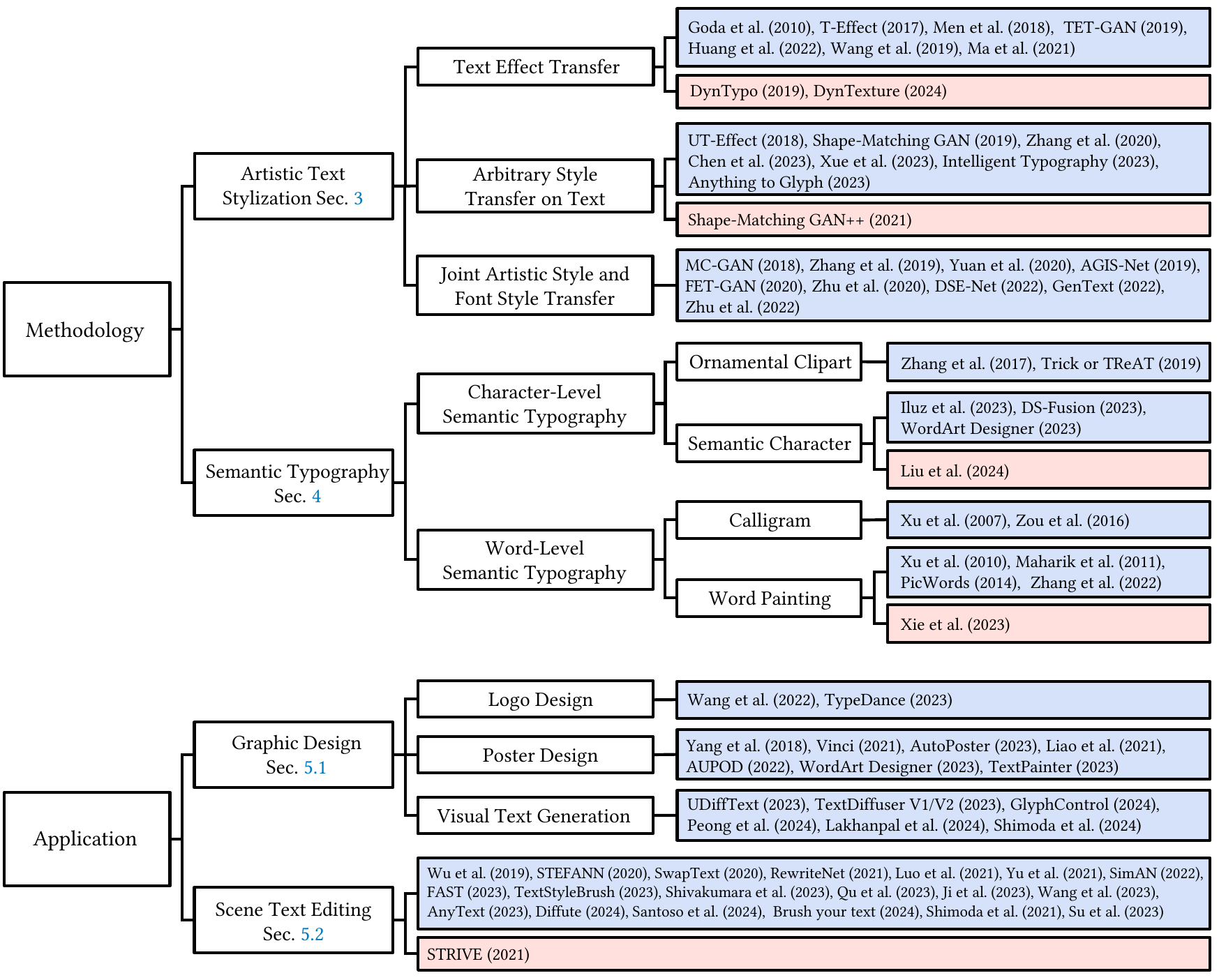}
    \caption{Taxonomy of the representative artistic text generation methods and applications. The blue background and red background indicate
    methods for static and dynamic artistic text generation, respectively.}
    \label{fig:taxonomy}
\end{figure}
\begin{itemize}
    \item \textbf{Text Effect Animation}. Text effect animation focuses on studying the transfer of dynamic text effects onto the static text. In most cases, the text itself remains stationary.
    Sometimes, motion effects such as appearing, moving, scaling, or disappearing can be also considered as part of the text effects. In such a case, the output $T'$ engages both text motions and animated visual effects.
    \item \textbf{Text Shape Animation}. Text shape animation primarily focuses on how to animate the semantic typography to resemble the motion of the intended semantic content naturally. For example, animating the leg-kicking motion for the letter `L' in the word ``LEG''. This involves animating the shape of the text to bring it to life and can visually convey more abstract concepts.
\end{itemize}

We summarize the taxonomy of the representative artistic text generation methods in Figure~\ref{fig:taxonomy}.
Specifically, all methods can be roughly divided into artistic text stylization and semantic typography.
In the task of artistic text stylization, two kinds of styles are considered: text effects and arbitrary style, corresponding to text effect transfer and arbitrary style transfer on text. Based on the modality, artistic text stylization can be further divided into static and dynamic artistic text stylization. 
Meanwhile, in the task of semantic typography, character-level and word-level design are studied. 
Based on the modality, semantic typography can be further divided into static semantic typography and kinetic typography. 
In the following sections, we will detail the main ideas of the representative methods in the order of their categories.

\section{Artistic Text Stylization}
\label{sec3}

\begin{table} [t]
\caption{Summary of artistic text generation methods.}\vspace{-1mm}
\label{tb:control}
\centering
\scriptsize
\resizebox{\linewidth}{!}{
\begin{tabular}{l|c|c|l}
\toprule
\textbf{Method} & \textbf{Style type} & \textbf{Model type} & \textbf{Feature}  \\
\midrule
\multicolumn{4}{c}{\textit{Static Artistic Text Stylization}} \\
\midrule
Goda \etal~\cite{Goda2010TextureTB} & calligraphy  & stroke-based & ink texture synthesis along strokes \\
T-Effect~\cite{yang2017awesome} & text effects  & patch-based & distribution-aware text effect prior \\
Men \etal~\cite{men2018common} & text effects & patch-based & versatile interactive texture transfer \\
TET-GAN~\cite{yang2019tet} & text effects  & GAN-based & style-glyph disentanglement\\
Huang \etal~\cite{huang2022artistic} & text effects  & GAN-based & simplified TET-GAN~\cite{yang2019tet} \\
Wang \etal~\cite{wang2019typography} & text effects  & GAN-based & decorative element (decor) transfer \\
Ma \etal~\cite{Ma2021TextST} & text effects  & GAN-based & decor transfer on Chinese characters \\
UT-Effect~\cite{yang2018context} & arbitrary style & patch-based & structure transfer \& texture transfer\\
Shape-Matching GAN~\cite{yang2019shapematching} & arbitrary style & GAN-based & one-shot learning; style degree control\\
Zhang \etal~\cite{zhang2020improving} & arbitrary style & GAN-based & clean edges by erosion and dilation \\
Chen \etal~\cite{chen2023style} & arbitrary style & GAN-based & clean edges by erosion and dilation \\
Xue \etal~\cite{xue2023art} & arbitrary style & GAN-based & train a network to generate data \\
Intelligent Typography~\cite{Mao2023IntelligentTA} & arbitrary style & GAN-based & coarse-to-fine complex style transfer \\
Anything to Glyph~\cite{wang2023anything} & arbitrary style & diffusion-based & place objects according to the glyph \\
MC-GAN~\cite{azadi2018multi}& text effects \& font & GAN-based & end-to-end stack network \\
Zhang \etal~\cite{zhang2019neural} & text effects \& font & GAN-based & coarse-to-fine cascaded stack network\\
Yuan \etal~\cite{yuan2020art} & text effects \& font & GAN-based & text edge and skeleton as auxiliary input \\
AGIS-Net~\cite{gao2019artistic} & text effects \& font & GAN-based & two parallel encoder-decoder branches \\
FET-GAN~\cite{li2020fet} & text effects \& font & GAN-based & AdaIN-based text style modelling\\
Zhu \etal~\cite{zhu2020few} & text effects \& font & GAN-based & weighted style representation \\
DSE-Net~\cite{li2022dse} & text effects \& font & GAN-based & effect-font-glyph disentanglement \\
GenText~\cite{huang2022gentext} & text effects \& font & GAN-based & multi-task end-to-end training \\
Zhu \etal~\cite{zhu2022text} & text effects \& font & GAN-based & effect-font-glyph disentanglement \\
\midrule
\multicolumn{4}{c}{\textit{Dynamic Artistic Text Stylization}} \\
\midrule
DynTypo~\cite{men2019dyntypo} & text effects & patch-based & global NNF search across frames \\
DynTexture~\cite{pu2024dynamic} & text effects & patch \& Transformer & long-distance dependency modeling \\
Shape-Matching GAN++~\cite{yang2021shapematching++} & arbitrary style & GAN-based & spatial-temporal structural mappings \\
\midrule
\multicolumn{4}{c}{\textit{Static Semantic Typography}} \\
\midrule
Zhang \etal~\cite{zhang2017synthesizing} & ornamental clipart & retrieval-based & joint semantic and shape matching \\
Trick or TReAT ~\cite{tendulkar2019trick} & ornamental clipart & retrieval-based & unsupervised autoencoder matching \\
Iluz \etal~\cite{iluz2023word} & semantic character &  diffusion-based & vector glyph shape deformation\\
DS-Fusion~\cite{tanveer2023ds} & semantic character & diffusion-based & raster semantic feature enhancement \\
WordArt Designer~\cite{he2023wordart} & semantic character & LLM \& diffusion & user-controllable artistic design \\

Xu \etal~\cite{xu2007calligraphic} & calligram & warp-based & shape adaptive text warping \\
Zou \etal~\cite{zou2016legible} & calligram & warp-based & legibility enhanced calligram \\

Xu \etal~\cite{xu2010structure} & word painting & structure-based & structural ASCII art generation \\
Maharik \etal~\cite{Maharik2011Micrography} & word painting & vector field-based & adaptive text layout synthesis \\
PicWords~\cite{hu2014picwords} & word painting & warp-based & keyword semantic priority ranking \\
Zhang \etal~\cite{zhang2022creating} & word painting & vector field \& SVM & visual saliency for aesthetic optimization \\
\midrule
\multicolumn{4}{c}{\textit{Kinetic Typography}} \\
\midrule
Liu \etal~\cite{liu2024dynamic} & semantic character & diffusion-based & character deformation and animation \\
Xie \etal~\cite{xie2023creating} & word cloud & frame-based & emotional word cloud animation\\
\bottomrule 
\end{tabular}}
\end{table}

Artistic text stylization involves the rendering of various visual effects, such as colors, shadows, outlines, gradients, textures, and embellishments, to enhance the aesthetics of text. 
First of all, text stylization pertains to image stylization, focusing on how to apply image stylization techniques to the specific content of text images. 
In general, the development of artistic text stylization has largely followed the trajectory of image stylization techniques. Chronologically, image stylization has moved from traditional texture synthesis to image translation leveraging deep neural networks, and most recently, to more sophisticated text-to-image techniques utilizing diffusion models. 
Concurrently, artistic text stylization approaches have incorporated specific designs into the above techniques to better transfer text effects and maintain the legibility of text.
This section presents an overview of existing methods, highlighting their specific designs, strengths, and limitations.

\subsection{Static artistic text stylization}

\subsubsection{Text effect transfer}

\begin{figure}
    \centering
    \includegraphics[width=1\linewidth]{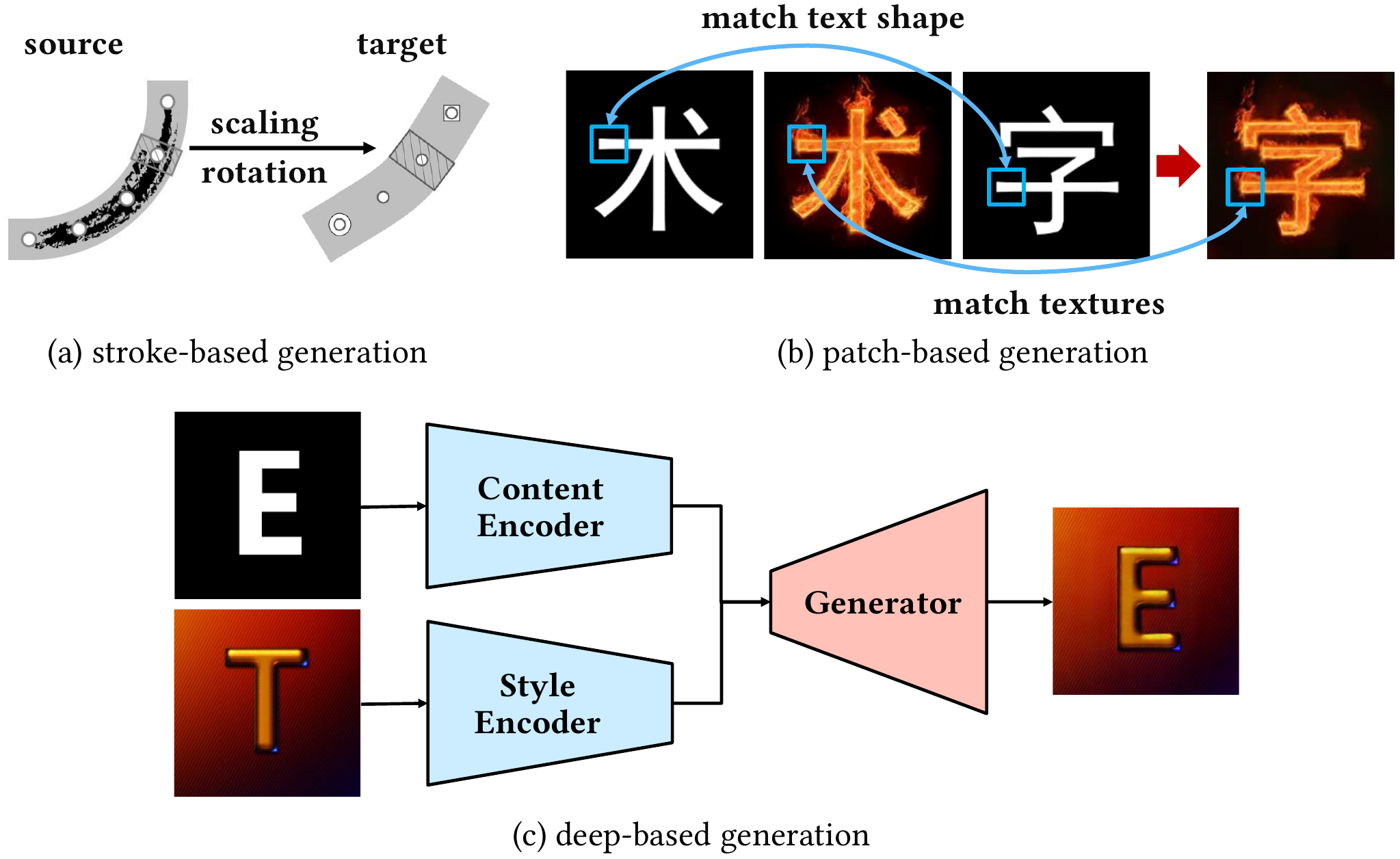}
    \caption{Different frameworks for artistic text stylization. Image credits: Goda \etal~\cite{Goda2010TextureTB}, T-Effect~\cite{yang2017awesome}, TET-GAN~\cite{yang2019tet}.}
    \label{fig:framework1}
\end{figure}

\textbf{Stroke-based text effect transfer.} Early image stylization methods primarily simulated how artists paint on the canvas: modeling brush strokes and synthesizing them onto a digital canvas~\cite{litwinowicz1997processing,hertzmann1998painterly,hertzmann2001paint}. 
Correspondingly, early text stylization methods focused on stylizing the strokes of characters, with the most representative research being calligraphy. 
In calligraphy, brush strokes vary in depth, pressure, and ink textures. 
Early studies~\cite{xu2005automatic} concentrated on synthesizing ink textures onto brush strokes. Directly applying texture synthesis technology~\cite{jacobs2001image} to characters often resulted in ink directions that did not align with the stroke directions. 
To address this, as illustrated in Figure~\ref{fig:framework1}(a), Goda \etal~\cite{Goda2010TextureTB} propose to synthesize ink textures along the edges of characters by finding textures with lengths and directions that match the strokes, yielding more natural results.

\begin{figure}[htbp]
    \centering
    \includegraphics[width=0.98\linewidth]{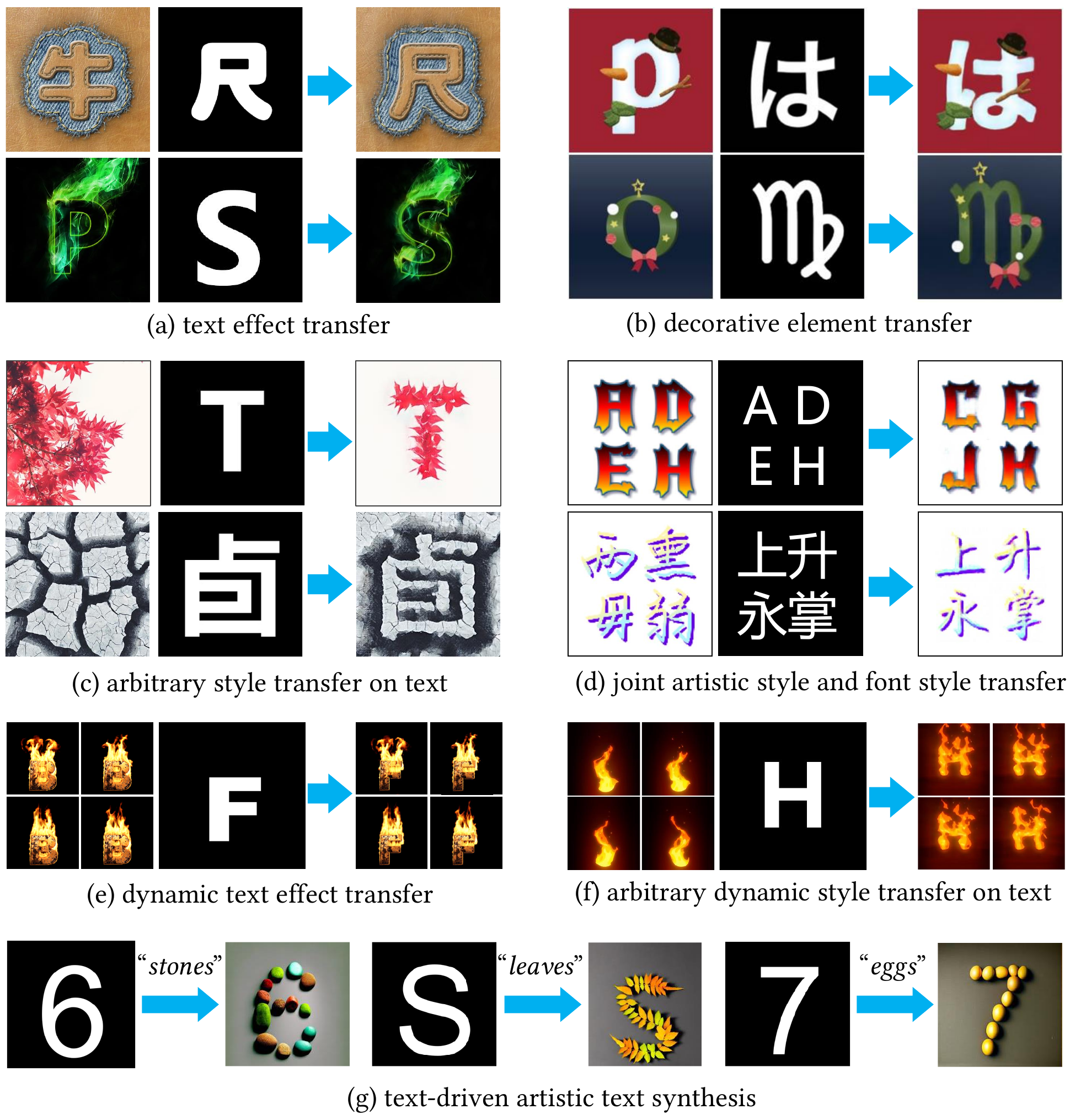}
    \caption{Artistic text stylization results of representative methods. Image credits: (a) T-Effect~\cite{yang2017awesome}. (b) Wang~\etal~\cite{wang2019typography}. (c) ShapeMatching GAN~\cite{yang2019shapematching}. (d) AGIS-Net~\cite{gao2019artistic}. (e) DynTypo~\cite{men2019dyntypo}. (f) ShapeMatchingGAN++~\cite{yang2021shapematching++}. (g) Anything to Glyph~\cite{wang2023anything}.}
    \label{fig:sec3_result}
\end{figure}

\textbf{Patch-based text effect transfer.} 
Calligraphy represents only a small portion of text effects, characterized by relatively monochrome colors. 
To handle more diverse and colorful text effects, 
T-Effect~\cite{yang2017awesome} introduces the first style transfer method specifically for general text effects, such as shadows, outlines, gradients, and textures. 
In particular, T-Effect defines a supervised text effect transfer problem: $S:S'::T:T'$~\cite{jacobs2001image}, where the source text effect $S'$ in addition to its corresponding plain text $S$ are required. The algorithms learn the transformation between them and then apply it to the target text $T$ to synthesize the result $T'$.
Methodologically, T-Effect builds on the texture synthesis method of Wexler \etal~\cite{Space-time} and its variants~\cite{ImageMelding} using random search and propagation as in PatchMatch~\cite{PatchMatch,GeneralizedPatchMatch}.
As illustrated in Figure~\ref{fig:framework1}(b), the basic texture synthesis process is to match patches between $S'$ and $T'$ (to match textures), as well as $S$ and $T$ (to match text shape), and to update each patch in $T'$ with its best-matched patch in $S'$. The process iteratively matches and updates patches until convergence. 
To extend this general texture synthesis to text effect, T-Effect analyzes real-world text effect images and summarizes a novel text effect prior: 
there is a high correlation between patch patterns (\ie, color and scale) and their distances to text skeletons in high-quality text effects. This is because the artists commonly adjust the effects based on text shapes for readability.
Based on this, T-Effect makes the following two modifications:
\begin{itemize}
    \item Scale-adaptive matching: Encourages the patch to be matched at their optical scale based on their distance to the text skeletons. This could preserve both coarse structures and texture details.
    \item Distribution-aware matching: Encourages the text effects of $T'$ to share similar distribution with $S'$. This could effectively realize spatial-aware style transfer. 
\end{itemize}
Besides, the T-Effect further considers avoiding texture over-repetition for more natural synthesis. By incorporating priors specific to text effects into the patch match approach, T-Effect enables more plausible artistic text generation as in Figure~\ref{fig:sec3_result}(a). 

Men \etal~\cite{men2018common} extend T-Effect to more general interactive texture transfer applications, where $S$ and $T$ can be general semantic maps. Since semantic maps have large flat regions that provide few cues for valid matching, this method introduces novel structure guidance. 
Men \etal~find that it is easier to build correspondences near the contours of the semantic maps. Based on the matched contour key points, the method propagates the structure guidance into inner flat regions to build rough correspondences, based on which textures are synthesized. Thus, this method can better transfer inner textures than T-Effect.

However, the patch-based method requires iterative patch matching, resulting in a generation time of approximately one minute per image, which is not sufficiently efficient.

\textbf{Deep-based text effect transfer.}
Entering the era of deep learning, researchers investigate the way of artistic text generation via data-driven learning. 
TET-GAN~\cite{yang2019tet} is one of the first deep methods and enables real-time artistic text generation after training. TET-GAN builds a text effect dataset with paired plain text images and the corresponding artistic text images. Then, the problem becomes to learn a supervised image-to-image translation as in pix2pix~\cite{isola2017image}.
However, directly applying pix2pix~\cite{isola2017image} can only learn one style at a time, which is less efficient in practice.
To this end, TET-GAN introduces the idea of style-content distentanglement~\cite{lee2018diverse,huang2018multimodal,royer2020xgan} into text effect transfer with separate content and style encoders as shown in Figure~\ref{fig:framework1}(c). Specifically, TET-GAN builds a multi-task framework trained with three tasks:
\begin{itemize}
    \item Glyph reconstruction: The network is trained to reconstruct the plain text image $T$ so that it learns the glyph features.
    \item Artistic text destylization: The network is trained to infer the glyph features from the artistic text images $S'$ so that it learns to disentangle the content representation.
    \item Artistic text stylization: The network is tasked to transfer the text effects of $S'$ onto $T$, obtaining the output that approaches the ground truth $T'$. The network will learn to disentangle the style representation and combine it with the content representation for style transfer. 
\end{itemize}
With this design, TET-GAN can learn hundreds of text effects in a single network. To further improve the practicality, it proposes a few-shot text effect fine-tuning strategy to efficiently extend the model to new text effects with only several and even one reference text effect image available.

Training on multiple tasks boosts the versatility of the model, which also increases the model complexity. Huang \etal~\cite{huang2022artistic} find that training a pix2pix network to map a channel-wise concatenated input of $S$, $T$ and $S'$ to the output $T'$ is enough for multi-style transfer. This is useful when complicated functions like style extension and interpolation featured in TET-GAN are not required. 

The aforementioned methods mainly treat text effects as a whole. However, some text effects have exquisite decor that needs special consideration. These decorative elements usually have very different styles from the base text effects. To address this problem, Wang \etal~\cite{wang2019typography} propose to learn to separate, transfer and recombine the decors and the base text effects. 
The framework contains a network for decorative element segmentation, text effect transfer, and structure-based decor recomposition. 
During recomposition, the decorative elements are divided into two classes based on their importance. Insignificant elements are repeatable and randomly scattered on the text, while the significant elements are placed based on their spatial distributions in the original $S'$. The method can produce professional artistic typography on English letters and simple symbols as shown in Figure~\ref{fig:sec3_result}(b). Ma \etal~\cite{Ma2021TextST} further extend this work to complex Chinese characters.  

Although the aforementioned text effect transfer methods have achieved great success in synthesizing professionally artistic text images as in Figure~\ref{fig:sec3_result}(a)(b), they can only mimic the well-defined reference text effects. When it comes to more general arbitrary reference style images like the fire, water, and leaves (\eg, Figure~\ref{fig:sec3_result}(c)(f)), as in common image style transfer tasks, these methods will fail. To handle these cases, researchers have paid attention to arbitrary style transfer on text.

\subsubsection{Arbitrary style transfer on text}

Text effect transfer is a well-defined problem with inputs $S$, $S'$, and $T$, and the ground truth output $T'$. However, for arbitrary style transfer on text, we do not have the plain text version for $S'$, and usually the ground truth output $T'$ is not available, posing an unsupervised image-to-image translation problem for researchers. 
Furthermore, unlike text effects, there is a significant visual discrepancy between the plain text and the colorful style images. 
Therefore, this task is more challenging. The key is to find appropriate correspondences between the two distant domains. 

\textbf{Patch-based style transfer on text.} 
UT-Effect~\cite{yang2018context} is one of the earliest methods to transfer arbitrary style onto text. To build valid correspondences between $S'$ and $T$, it proposes extracting a binary mask $S$ from $S'$ based on texture removal~\cite{xu2012structure}, super-pixel extraction~\cite{achanta2012slic} and clustering. 
The region with higher saliency is set as the foreground corresponding to the text region in $T$ while the remaining region is the background. Then the unsupervised problem becomes a supervised problem as proposed in T-Effect~\cite{yang2017awesome}. However, there is still obvious structure discrepancy between $S$ and $T$. For example, $S$ might be maple leaves of different shapes, while $T$ is the plain text with rigid edges. Directly synthesizing maple textures into $T$ will result in unnatural maple boundaries. To solve this problem, UT-Effect proposes a two-stage style transfer framework: 
\begin{itemize}
    \item Structure transfer: UT-Effect transfers the structure styles of $S$ onto $T$, obtaining $\hat{T}$ that shares similar boundaries with $S$ while maintaining the glyph of $T$. To achieve this, UT-Effect proposes a legibility-preserving structure transfer method, which uses a patch-based shape synthesis technique~\cite{rosenberger2009layered} to adjust the shape of the stroke ends while preserving the shape of the stroke trunk for legibility.
    \item Texture transfer: For the translation problem $S:S'::\hat{T}:\hat{T}'$, UT-Effect leverages the patch-based texture synthesis technique of T-Effect~\cite{yang2017awesome} and introduces a new saliency term to guide patch matching.  The saliency term encourages pixels inside the text to find salient textures for synthesis and keeps the background less salient, which makes the artistic text stand out from the background.
\end{itemize}

\textbf{GAN-based style transfer on text.} 
Applying deep learning to arbitrary style transfer on text is challenging since there is generally no large-scale ground truth data for model training. To solve this problem, Shape-Matching GAN~\cite{yang2019shapematching} proposes a one-shot bidirectional shape-matching framework to establish an effective glyph-style mapping at various deformation levels without paired ground truth. It includes two stages:
\begin{itemize}
    \item Backward transfer: After extracting the structure map $S$ from the style image $S'$ as in UT-Effect~\cite{yang2018context}, the first stage backward transfers the shape style of the text to the structure map, obtaining its sketchy or simplified version $\bar{S}$, whose contour style is similar to the plain text.
    \item Forward transfer: The second stage learns the forward mapping from the sketchy structure map $\bar{S}$ to the original structure map $S$, and further to the original style image $S'$. The network learns to characterize the shape and texture features of the style image in the training phase and transfers these features to the target text in the testing phase.
\end{itemize}
To enable training on a single style image, Shape-Matching GAN randomly crops $\bar{S}$, $S$, and $S'$ into sub-image pairs to obtain enough data. 
Another key contribution of Shape-Matching GAN is the style degree control mechanism. Specifically, there is a trade-off between legibility and artistry: The stylistic degree or shape deformations of a glyph need manipulation to resemble the style subject in $S'$, while the glyph legibility needs to be maintained so that the stylized text is still recognizable. People's diverse preferences make it difficult to define an optimal style degree. Shape-Matching GAN introduces an extra parameter $\ell$ to control the style degree freely to allow users to select the most desired one. During backward transfer, $\ell$ is used to control the simplification degree of $\bar{S}$ to build multi-degree paired data, thus during forward transfer, the model will learn multi-degree structure transfer conditioned on $\ell$. 

Based on Shape-Matching GAN~\cite{yang2019shapematching}, several improvements are proposed. Chen \etal~\cite{chen2023style} and Zhang \etal~\cite{zhang2020improving} use erosion and dilation to remove the unnecessary artifacts along the contour to generate cleaner structure transfer results. Zhang \etal~\cite{zhang2020improving} utilize multi-resolution style images borrowed from pyramid features~\cite{he2015spatial} for better texture transfer. Xue \etal~\cite{xue2023art} directly train a dataset generation network to synthesize various paired data to overcome the problem of limited data. 

Intelligent Typography~\cite{Mao2023IntelligentTA} finds that it is hard for a $1\times$ network to learn a robust pixel-to-pixel level relationship from a single style image $S'$ due to over-fitting. To overcome this issue in ShapeMatching GAN-based methods, Intelligent Typography develops a novel $2\times$ magnification network to smartly convert the problem of complex style transfer into texture expansion and super-resolution, which relieves the pressure of one-shot learning dramatically. To better transfer the style effects with relatively complex texture and structure, Intelligent Typography proposes a coarse-to-fine framework with two stages: prototype generation and detail refinement. Prototype generation generates a coarse-level stylized prototype from the given mask and tailored texture with the $2\times$ magnification network.
Then, a structure network and a texture network are proposed to refine the details of the prototype. With the above design, Intelligent Typography can generate exquisite images with vivid artistic text details and clear backgrounds. 

\textbf{Diffusion-based style transfer on text.} 
Recently, the increase in data scale and model representation capabilities has ultimately led to the emergence of large diffusion models~\cite{rombach2022high}. Diffusion models bring new opportunities for artistic text generation. Diffusion models exhibit unprecedented expressive power, providing diverse style support and interactive generation control with the help of large language models~\cite{radford2021learning}. 
Anything to Glyph~\cite{wang2023anything} is one of the recent diffusion-based, text-driven artistic text synthesis methods. It leverages the generative power of pre-trained Stable Diffusion~\cite{rombach2022high} and segmentation models~\cite{luddecke2022image} to generate a paired dataset and train a diffusion model called Position Predictor to predict an object's position mask $S$ in $S'$. Then, during testing, Denoising Score Matching~\cite{vincent2011connection} is applied to update the latent code of the Position Predictor so that its denoising result is similar to the target text shape $T$ while maintaining the consistency with the prompt. 
Given the updated latent code, which represents an initial structure transfer result (like $\hat{T}$ in UT-Effect~\cite{yang2018context}), pre-trained Stable Diffusion is used to synthesize textures onto it under the guidance of the prompt. 
Anything to Glyph is especially good at generating artistic text images composed of multiple instances of objects specified by the prompt such as stones, leaves, and eggs as shown in Figure~\ref{fig:sec3_result}(g). 

\subsubsection{Joint artistic style and font style transfer}

The font is an important style of text complementary to the text effects. 
Different from text effects, large-scale text images with different fonts can be easily generated, which is suitable for deep learning. 
With the rapid development of deep generative models, font generation~\cite{atarsaikhan2017neural,zhang2018separating,xie2021dg,park2021few,wang2023cf} has become a hot topic and made great progress. 
In addition to the large-scale supervised learning, researchers focus on investigating more challenging few-shot font generation~\cite{park2021few,wang2023cf} and handwriting generation~\cite{Haines2016My,Lian2016Automatic}.
Font generation can effectively design and produce new typefaces that can be used in various digital and print media. The generation process can automate the creation of font styles, weights, and variations, making it easier to produce large font families with consistent design characteristics. 
This section will briefly review representative approaches that combine font transfer and text effect transfer. High-quality joint transfer results are shown in Figure~\ref{fig:sec3_result}(d)

MC-GAN~\cite{azadi2018multi} is one of the first deep-based few-shot joint font and text effect transfer models. It contains a glyph network for font generation and an ornamentation network for text effect transfer. MC-GAN stacks these two networks and jointly trains them to realize an end-to-end solution. Specifically, the glyph network is pre-trained on a large-scale dataset so that it can predict the coarse glyph shapes of the missing English letters from a few stylized English letter examples. The full model with two networks is then fine-tuned on the stylized examples to learn to refine and stylize coarse glyph shapes into clean and well-designed letters.

Zhang \etal~\cite{zhang2019neural} further divide the glyph network of MC-GAN~\cite{azadi2018multi} into a coarse-level image-to-image translation network and a fine-level stack network. In terms of text effect transfer, it utilizes the widely-used Neural Style Transfer~\cite{gatys2016image}. 
On the other hand, Yuan \etal~\cite{yuan2020art} find that using edge and skeleton information of $T$ as auxiliary input of the glyph network could better infer the shape of the font. 

AGIS-Net~\cite{gao2019artistic} proposes to disentangle the content and style representations with two encoders and two decoders. Two encoders learn to extract the content feature and style feature, respectively. Then two collaborative decoders generate the glyph shape image and final artistic text image simultaneously based on the extract features. The disentanglement ensures a few-shot multi-content and multi-style generation.

Inspired by the AdaIN-based style representation~\cite{huang2017real}, 
FET-GAN~\cite{li2020fet} views font style and text effect style as a whole and models them with channel-wise means and standard deviations as in~\cite{huang2017real}. The content feature extracted from the source image is stylized through AdaIN operation and further decoded to obtain the final stylized artistic text image. 

Zhu \etal~\cite{zhu2020few} also propose a few-shot end-to-end framework that extracts and combines content and style representations for joint font and text effect transfer. The key idea is that text content and style
are not fully independent. Therefore, it calculates the similarity between the reference glyph and the target glyph to assign weights for the style features of each referenced glyph. Then, the weighted style representation and the content representation are fused to generate the final artistic text image.

DSE-Net~\cite{li2022dse} argues that treating font style and text effect style as a whole would limit the transfer of complex styles. It presents a disentangled style encoding network, with three different networks to extract the font feature, text effect feature, and glyph content feature, respectively. Finally, a cross-layer fusion mechanism is proposed to fuse the features adaptively to generate the final output. 

GenText~\cite{huang2022gentext} extends TET-GAN~\cite{yang2019tet} with an extra task of font transfer. 
It treats the plain text as an artistic text image with special text effects, thus unifying the destylization and stylization tasks within a single network. Then, GenText uses an encoder network to extract the content code and the style code from artistic text images, a font transfer network to fuse the content code of the plain text and the style code of the plain text with reference font for font transfer, and a text effect transfer network to fuse the content code of the plain text and the style code of the artistic text for stylization and destylization. Then, artistic text can be generated by sequentially performing font transfer and stylization.

Zhu \etal~\cite{zhu2022text} propose an artistic text style transfer model based on multi-factor disentanglement and mixture. The model contains three encoders to extract the text effect, font, and glyph representations, and further employs adversarial training and one-factor swap training strategies to disentangle the three representations. The disentanglement enables several tasks of font transfer, text effect transfer, joint transfer, and style removal.

\subsection{Dynamic artistic text stylization}

Compared to static artistic text, dynamic artistic text is more attractive and is widely used in a variety of media such as films, advertisements, and video clips. 
While static artistic text stylization~\cite{yang2017awesome,yang2018context,yang2019tet,yang2019shapematching} and general video image style transfer~\cite{ruder2016artistic,gupta2017characterizing,huang2017real,chen2017coherent,wang2020Compound} have been extensively studied, 
a few approaches~\cite{men2019dyntypo,pu2024dynamic,yang2021shapematching++} have studied dynamic artistic text stylization, which is reviewed in this section. 

Different from general video style transfer that migrates static style from a style image onto a content video, dynamic artistic text stylization aims to transfer dynamic style from a style video $\mathbf{S}'=\{S'_1, S'_2, ..., S'_N\}$ onto a text image $T$ as illustrated in Figure~\ref{fig:sec3_result}(e)(f), where $N$ is the total frame number. Therefore, it is not straightforward to apply the optical flow guidance~\cite{ruder2016artistic,gupta2017characterizing,huang2017real,chen2017coherent} widely used in video style transfer to dynamic artistic text generation. Instead, dynamic artistic text stylization approaches mainly focus on spatial-temporal style representation modeling and transfer. 

\subsubsection{Dynamic text effect transfer}

DynTypo~\cite{men2019dyntypo} extends the NNF search of PatchMatch~\cite{PatchMatch} to the spatial-temporal domain.
Instead of searching the Nearest-neighbor Field (NNF) for text effect synthesis in a frame-by-frame manner, the main idea of DynTypo is to simultaneously optimize the text effect coherence across all frames to find a common NNF for all temporal frames. 
Specifically, DynTypo stacks patches at the same position but across an entire video into a patch cube and matches patches at a cube level. After matching, the entire cubes are directly used to synthesize the output video.
DynTypo first detects keyframes based on the intensity of text effect dynamics and then limits the procedure of cube matching within the keyframes to maintain both spatial and temporal consistencies. 
DynTypo further combines PatchMatch with Simulated Annealing, to add more priority to the patches near the text contours. 
With the above designs, DynTypo achieves impressive results (\eg, Figure~\ref{fig:sec3_result}(e)) in dynamic text effect transfer.

However, it is hard for a single global NNF to transfer complex dynamic text effects such as moving samples that shift across source videos. 
To better model the inter-frame correlation and transfer complex dynamic text effects, 
DynTexture~\cite{pu2024dynamic} proposes to combine PatchMatch~\cite{PatchMatch} with the advanced Transformers~\cite{vaswani2017attention}. 
Specifically, DynTexture decomposes the dynamic text effect transfer task into two stages.
\begin{itemize}
    \item First frame generation: DynTexture adopts patch-based text effect transfer~\cite{yang2017awesome} with distance map guidance~\cite{men2019dyntypo} to generate the first frame.
    \item Full video generation: The synthesized first frame is decomposed into structure-agnostic patches, which are then encoded to tokens.  
    Then, Transformers~\cite{vaswani2017attention} equipped with VQ-VAE~\cite{van2017neural} are exploited to predict the discretized token sequences, leveraging its high capability of capturing the long-distance dependencies between frames. All predicted patches are assembled into each frame by a Gaussian weighted average merging strategy to obtain the final full video result.
\end{itemize}

\subsubsection{Arbitrary dynamic style transfer on text}

In the scope of arbitrary style transfer, 
Shape-Matching GAN++~\cite{yang2021shapematching++} extends Shape-Matching GAN~\cite{yang2019shapematching} to video domains. 
Shape-Matching GAN learns the forward mapping from the sketchy structure map $\bar{S}$ to the original structure map $S$ in a patch level. 
In Shape-Matching GAN++, to learn spatial-temporal shape mappings, patches across consecutive frames are learned together, just as the patch cude in DynTypo~\cite{men2019dyntypo}, so that the model could learn inter-frame motion patterns. 
To generate long-term motions, Shape-Matching GAN++ divides the full video into multiple $K$-frame video clips and focuses on the short-term motion patterns of these video clips. 
It defines the short-term motion pattern as the shape dynamics between the frames within a video clip. During training, the model learns the shape mappings between the first $K-1$ frames and the last frame of a video clip, which is actually a frame prediction task. 
During testing, the model repeatedly predicts the next frame based on previous frames and propagates the short-term motion patterns to achieve long-term motion patterns. Figure~\ref{fig:sec3_result}(f) presents an example of the dynamic style transfer result by ShapeMatching GAN++.

\section{Semantic Typography}
\label{sec4}

Semantic Typography focuses on transforming the text shape to visually represent specific objects, themes, or concepts. This section investigates different techniques, including shape morphing, deformation, and adaptive typography, to realize this meaningful and visually appealing art form.

\subsection{Static semantic typography}

\subsubsection{Character-level semantic typography}

\begin{figure}[htbp]
    \centering
    \includegraphics[width=0.9\linewidth]{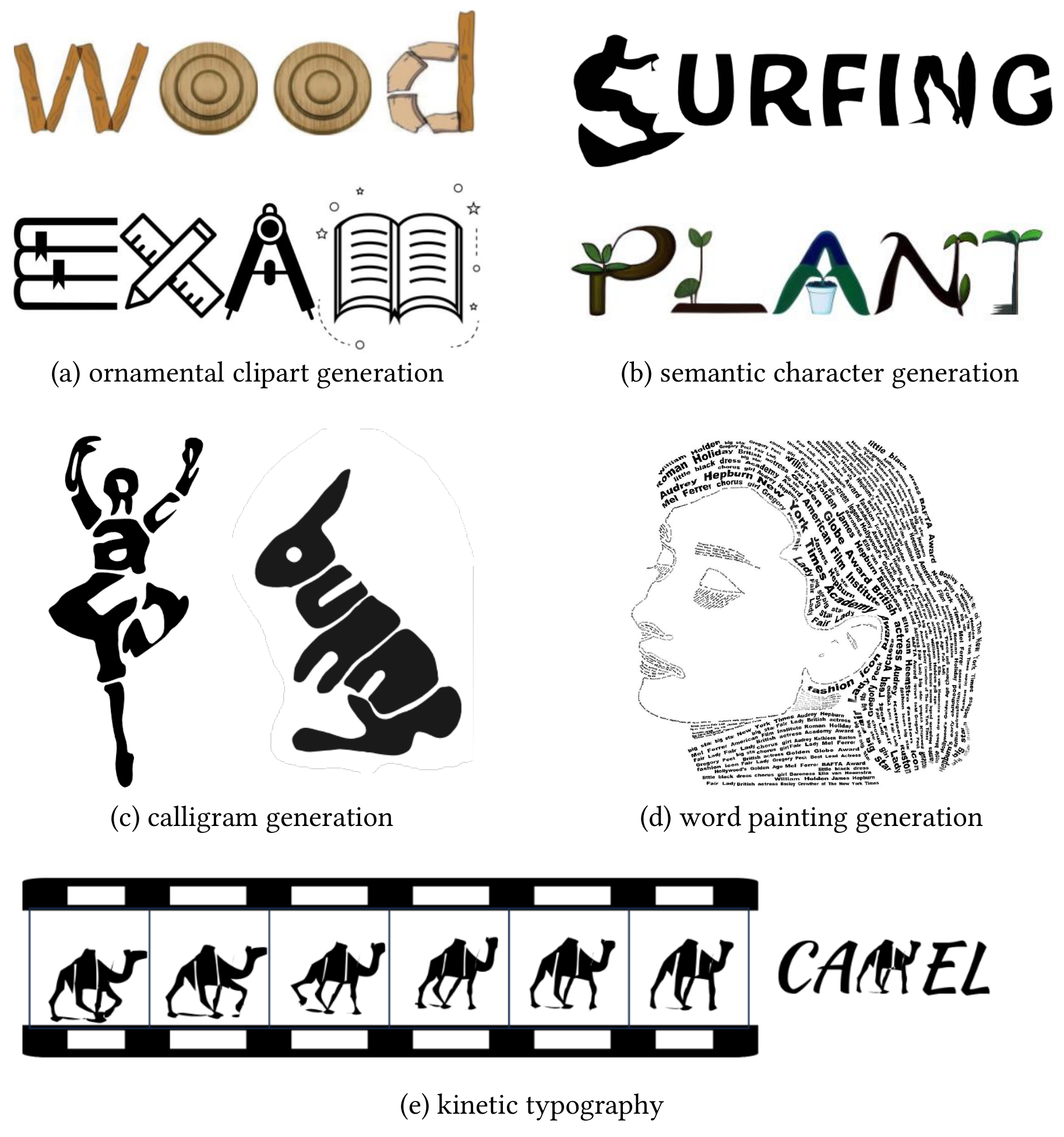}
    \caption{Semantic typography results of representative methods. Image credits: (a) Zhang \etal~\cite{zhang2017synthesizing}, Trick or TReAT ~\cite{tendulkar2019trick}. (b) Iluz \etal~\cite{iluz2023word}, DS-Fusion~\cite{tanveer2023ds}. (c)  Xu \etal~\cite{xu2007calligraphic}, Zou \etal~\cite{zou2016legible}. (d) Zhang \etal~\cite{zhang2022creating}. (e) Liu \etal~\cite{liu2024dynamic}. }
    \label{fig:sec4_result}
\end{figure}

\textbf{Ornamental Clipart Generation.}
Ornamental clipart generation aims to identify suitable images that correspond to the semantics of words and meticulously assemble them under the direction of glyph strokes as shown in Figure~\ref{fig:sec4_result}(a), which is considerably time-consuming when carried out manually. 
Zhang \etal~\cite{zhang2017synthesizing} present an automatic framework for creating such ornamental typefaces. 
Specifically, the framework features an interactive interface for the glyph stroke segment with scribbles from the user, thus fulfilling their personalized requirements. In terms of image selection, a semantic-shape similarity metric is established to concurrently account for both word semantics and stroke form. An optional structural optimization step based on gradient descent is implemented to yield results with enhanced glyph structure and aesthetic appeal.

Trick or TReAT ~\cite{tendulkar2019trick} is another retrieval-based method for ornamental clipart generation. It leverages cliparts to resemble the glyphs of characters to express the semantic features of words. To better retrieve the similarity between letters and cliparts, it trains an auto-encoder based on AlexNet~\cite{krizhevsky2017imagenet} in an unsupervised manner to automatically learn a hidden feature space. The corresponding cliparts will be retrieved based on the distance in the hidden space. Trick or TReAT generates results that have good legibility, semantics, and creativity. However, as with all retrieval-based methods, the diversity of the output is inherently dependent on the input data, which may introduce some limitations in the richness of the generated content.

\textbf{Semantic Character Generation.} 
With the development of generative models in recent years, diffusion models have demonstrated exceptional performance in image synthesis tasks~\cite{ramesh2021zero,ho2020denoising,rombach2022high,saharia2022photorealistic,ramesh2022hierarchical}.  The robust cross-modal understanding and rich semantic priors encapsulated make diffusion models highly adaptable for semantic typography. As shown in Figure~\ref{fig:sec4_result}(b), Iluz \etal~\cite{iluz2023word} pioneer the integration of a pre-trained Stable Diffusion model into the semantic character generation process. Focusing on the geometric transformation of glyphs, they employed the SDS loss~\cite{poole2022dreamfusion} and DiffVG to tailor diffusion models for the artistic vectorized glyph shape generation. Additionally, they incorporated the as conformal as possible (ACAP) loss~\cite{hormann2000mips} to regulate the extent of character deformation and established a tone preservation loss to maintain the structural integrity of the original glyph. This approach not only enhances the adaptability of diffusion models but also ensures that the generated characters retain their distinctive aesthetic and semantic qualities.

In contrast to the approach of Iluz \etal~\cite{iluz2023word}, DS-Fusion~\cite{tanveer2023ds} places a greater emphasis on texture and color features by directly processing glyph images in raster form. Guided by style prompts and glyph images, DS-Fusion integrates the glyph shape with style images derived from a pre-trained Stable Diffusion model~\cite{rombach2022high}, which is conditioned with style prompts. A CNN-based discriminator is employed to provide implicit supervision of the generation process by examining the feature maps of both the glyph shape and the generated image in latent space. It is noteworthy that DS-Fusion, when supplied with an entire word as a glyph reference, is also adept at producing word-level stylized results that exhibit harmonious character combinations and a vivid artistic expression.

WordArt Designer~\cite{he2023wordart} presents a user-controllable artistic semantic character design system with a large language model (LLM) engine~\cite{achiam2023gpt}. It contains three modules SemTypo, StyTypo, and TextTypo. SemTypo employs character parameterization and rasterization techniques akin to those used by Iluz \etal~\cite{iluz2023word}, to manipulate semantic features and deform selected parts of character strokes. StyTypo refines the smoothness and stylization details by leveraging the depth2image approach of latent diffusion models~\cite{rombach2022high}. In TextTypo, a ControlNet~\cite{zhang2023adding} is conditioned with Canny edges, depth maps, scribbles, and the original text image, enabling it to render textures that align with semantic styles and glyphs. The LLM engine generates structured text prompts based on user descriptions, feeding into the aforementioned models and thereby amplifying the creative diversity. 

\subsubsection{Word-level semantic typography}

\textbf{Calligram Generation.}
A calligram is a creative arrangement of words or letters that forms a visual image, conveying both meaning and aesthetics, shown in Figure~\ref{fig:sec4_result}(c).  Xu \etal~\cite{xu2007calligraphic} first introduce a warp-based method to integrate letters into the subject region of an image, thereby crafting calligrams that possess semantic features alongside logical glyph stroke deformation. This method employs an interactive approach to divide the container into subregions, which are subsequently filled by warping letters. While this approach can produce calligrams with coherent letter shape arrangements, it sometimes struggles with legibility. 

To address this issue, Zou \etal~\cite{zou2016legible} conduct a crowd-sourced study aimed at refining the guidance for glyph deformation to enhance letter legibility. They introduced a fully automatic method for generating letter layouts, aligning letters based on correspondence, and applying deformation.

\textbf{Word Painting Generation.}
Word painting represents a form of composite artwork, characterized by the adaptive fusion of visual structure and texture derived from a source image with semantic features extracted from a textual source, shown in Figure~\ref{fig:sec4_result}(d). Xu \etal~\cite{xu2010structure} formulate this task as approximating the primary structure of a reference image using ASCII characters. This method surpasses traditional tone-based techniques by capturing structural and semantic nuances through an innovative alignment-insensitive shape similarity metric. Maharik \etal~\cite{Maharik2011Micrography} introduce a technique for crafting micrographics, a distinctive variant of word painting that comprises minuscule letters. Departing from the conventional single horizontal text layout of ASCII art, this method emphasizes the design of low curvature and smooth vector fields devoid of singularities, facilitating the synthesis of word layouts that conform to regional shapes while maintaining high text readability. Additionally, a warping procedure for text height and width is proposed, enabling the adjustment of text shape to seamlessly integrate with the image contours.

PicWords~\cite{hu2014picwords} is an automatic word painting generation framework that employs non-photorealistic rendering (NPR) techniques. It processes the input image by segmenting the binary silhouette into distinct patches, each designed to encapsulate a keyword. By ranking the patches and keywords, this approach establishes a keyword-patch correspondence that serves to accentuate significant keywords, ensuring that key semantic information is effectively conveyed. Zhang \etal~\cite{zhang2022creating} introduce a method that utilizes a smooth vector field for patch segmentation, coupled with a Support Vector Machine (SVM)-based visual attention model to optimize the aesthetic arrangement of text. This visual attention model is trained to generate a saliency map for a given image, strategically positioning keywords that are closely related to the theme in areas that naturally draw the viewer's attention. Compared with prior work~\cite{Maharik2011Micrography}, this approach is capable of producing results that are imbued with more profound semantic information, enhancing both the visual appeal and the narrative depth of the generated word paintings.

\subsection{Kinetic typography}
Kinetic typography is a dynamic and expressive art form that combines motion graphics with typography to convey emotions, narratives, or messages in a visually engaging way. It involves animating text so that it moves, transforms, or interacts with other elements in a sequence or scene. Early works focus on exploring the interaction between dynamic forms and content to enrich emotional expression~\cite{Ford1997KineticTI,lee2006using}, with various kinetic typography system designs for animating text~\cite{lee2002kinetic,forlizzi2003kinedit,minakuchi2005automatic}. Wakey-Wakey~\cite{xie2023wakey} presents an automatic framework for aligning dynamic text motions with GIF animation. However, these methods lack effective measures to combine semantic features with textual dynamics.

Liu \etal~\cite{liu2024dynamic} introduce a novel approach to generate text animation with semantic features. In contrast to the work of Iluz \etal~\cite{iluz2023word}, they have developed an end-to-end model designed to mitigate conflicts with prior knowledge. As shown in Figure~\ref{fig:sec4_result}(e), this model leverages neural displacement fields and vector representations to deform letters, thereby conveying semantic meanings and rendering them in dynamic movements that respond to user prompts. Xie \etal~\cite{xie2023creating} propose a method to animate compact word clouds, expanding the scope from single-word animations to multiple words that express semantic emotions. 

\section{Applications}
\label{sec5}

\subsection{Graphic Design}

\begin{figure}[htbp]
    \centering
    \includegraphics[width=1\linewidth]{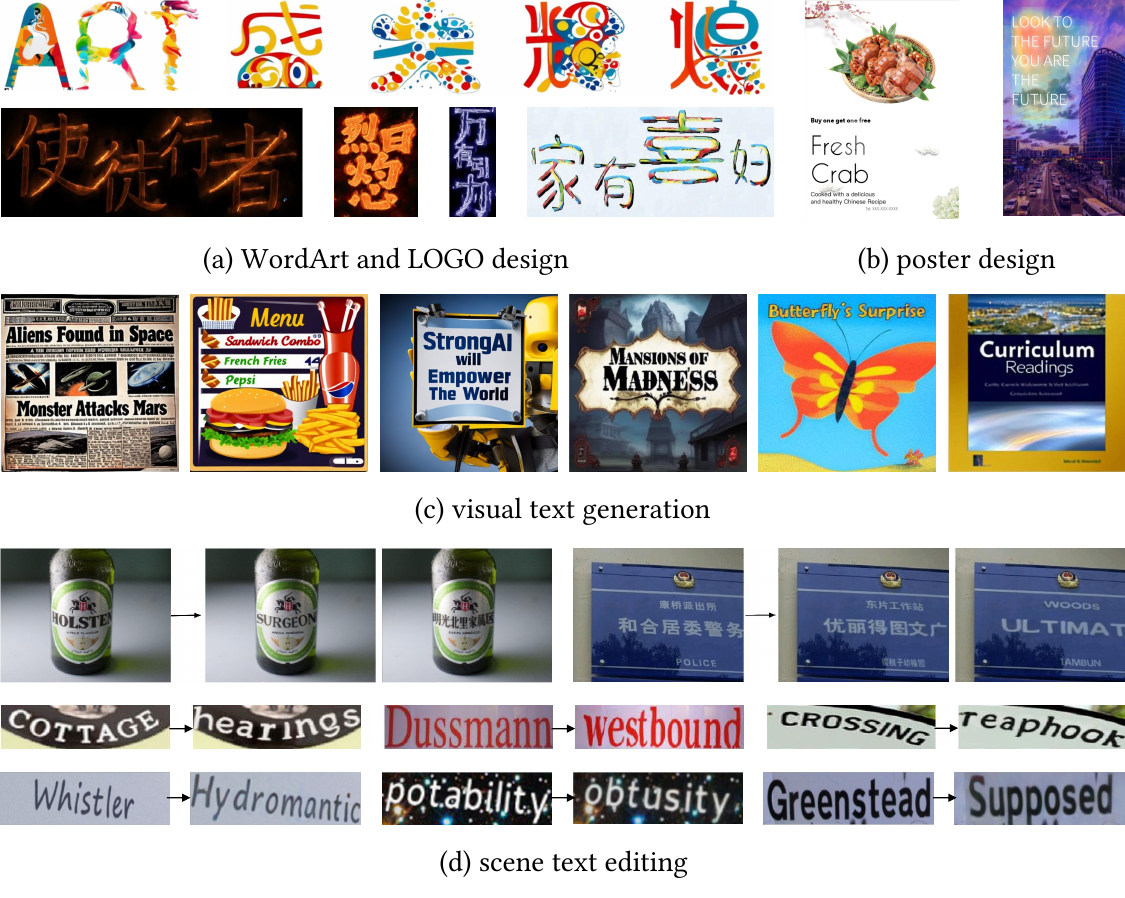}
    \caption{The examples of applications. (a) WordArt and LOGO design by WordArt Designer~\cite{he2023wordart}, Wang \etal~\cite{wang2022aesthetic}. (b) Poster design by Vinci~\cite{guo2021vinci}, AUPOD~\cite{huang2022aupod}. (c) Visual text generation by GlyphControl~\cite{yang2024glyphcontrol}, TextDiffuser-2~\cite{chen2023textdiffuser}. (d) Scene text editing by Qu \etal~\cite{qu2023exploring}, SwapText~\cite{yang2020swaptext}, FAST~\cite{das2023fast}.}
    \label{fig:applications}
\end{figure}

WordArt and LOGO design is a task involving the transformation of text input into semantically rich typography, incorporating various elements and layouts, shown in Figure~\ref{fig:applications}(a). Numerous studies have successfully addressed the generation of diverse WordArt and LOGO designs, utilizing a wide range of elements and layouts \cite{he2023wordart,wang2022aesthetic,xiao2024typedance}.

Similarly, poster design entails the arrangement and styling of provided text, subsequently generating the designed text on an input image, shown in Figure~\ref{fig:applications}(b). Several works~\cite{yang2018context2,guo2021vinci,lin2023autoposter,liao2021interactive,huang2022aupod,he2023wordart,gao2023textpainter} have achieved advancements in this area.

Visual text generation focuses on producing accurate and legible text within image generation, addressing the common limitation of image generation models, which often struggle to create readable text, shown in Figure~\ref{fig:applications}(c). Various studies have achieved notable success in generating images with clear and coherent text \cite{zhao2023udifftext,chen2024textdiffuser,chen2023textdiffuser,yang2024glyphcontrol,peong2024typographic,lakhanpal2024refining,shimoda2024towards}.

\subsection{Scene Text Editing}

Scene text editing has emerged as a notable research area in recent years, focusing on the transformation of text within a source image to align with a target reference text while preserving the original style and background, shown in Figure~\ref{fig:applications}(d). Numerous studies~\cite{wu2019editing,roy2020stefann,yang2020swaptext,lee2021rewritenet,luo2021separating,yu2021mask,luo2022siman,das2023fast,krishnan2023textstylebrush,shivakumara2023new,qu2023exploring} have proposed methodologies to generate high-quality text images despite varying backgrounds and fonts. 

Diffusion models have demonstrated significant potential in editing images of arbitrary topics, including scene text. To fully exploit this potential, several works~\cite{ji2023improving,wang2023letter,tuo2023anytext,chen2024diffute,santoso2024manipulating,zhang2024brush} have applied diffusion models to the task of scene text editing.

The subfield of scene-style text editing, addressed in some research \cite{shimoda2021rendering,su2023scene}, entails modifying specific attributes of text within an image while either preserving or altering its style. These attributes include rotation, font, color, and content, allowing for versatile text manipulations across various scene images.

Moreover, scene text editing can be extended to video applications \cite{subramanian2021strive}. Enhancing scene text editing in videos necessitates preserving geometric integrity, appearance consistency, and temporal coherence.

\begin{table*}[t]
\caption{Summary of the benchmark datasets for artistic text rendering and design}
\centering
\resizebox{\linewidth}{!}{
\begin{tabular}{cccc}
  \toprule
  \multirow{1}{*}{\textbf{Dataset}} &
  \multirow{1}{*}{\textbf{Type}} &
  \multirow{1}{*}{\textbf{Images}} &
  \multirow{1}{*}{\textbf{Feature}} \\
  \midrule
  \multirow{2}{*}{MC-GAN-Gray~\cite{azadi2018multi}}   &  & \multirow{2}{*}{260,000}  & \multirow{2}{*}{gray-scale English letter.} \\ 
  & \multirow{1}{*}{joint font \&} & & \\

  \multirow{2}{*}{MC-GAN-Color~\cite{azadi2018multi}}   & \multirow{1}{*}{text effects}  & \multirow{2}{*}{520,000}  & \multirow{2}{*}{colorful English letter.} \\ 
  & & & \\

  \midrule
  \multirow{2}{*}{AGIS-Net-C~\cite{gao2019artistic}}   &   & \multirow{2}{*}{1,571,940}  & \multirow{2}{*}{synthetic artistic Chinese characters.} \\ 
  & \multirow{1}{*}{joint font \&} & & \\

  \multirow{2}{*}{AGIS-Net-P~\cite{gao2019artistic}}   & \multirow{1}{*}{text effects}  & \multirow{2}{*}{256,410}  & \multirow{2}{*}{professional-designed artistic Chinese characters.} \\ 
  &  & & \\

  \midrule
  \multirow{2}{*}{TET-GAN~\cite{yang2019tet}}   & \multirow{2}{*}{text effects}  & \multirow{2}{*}{53,568}  & \multirow{1}{*}{64 text effects rendered on 775 Chinese characters,} \\ 
  & & & \multirow{1}{*}{52 English letters and 10 Arabic numerals.}\\

  \midrule
  \multirow{2}{*}{TextEffects-Decor~\cite{wang2019typography}}  & \multirow{2}{*}{text effects}  & \multirow{2}{*}{59,280}  & \multirow{1}{*}{64 text effects with decorative elements} \\ 
  & & & \multirow{1}{*}{rendered on 52 English letters of 19 fonts.}\\    

  \midrule
  \multirow{1}{*}{TE141K-E~\cite{yang2020te141k}}  &  & \multirow{1}{*}{66,196}  & \multirow{1}{*}{64 text effects on 52 English letters of 19 fonts.} \\ 

  \multirow{2}{*}{TE141K-C~\cite{yang2020te141k}}  & \multirow{2}{*}{text effects}  & \multirow{2}{*}{54,405}  & \multirow{1}{*}{65 text effects rendered on 775 Chinese characters,} \\ 
  & & & \multirow{1}{*}{52 English letters and 10 Arabic numerals.}\\   

  \multirow{2}{*}{TE141K-S~\cite{yang2020te141k}}  &  & \multirow{2}{*}{20,480}  & \multirow{1}{*}{20 text effects rendered on 56 special symbols,} \\ 
  & & & \multirow{1}{*}{and 968 letters in Japanese, Russian, \etc.}\\   

  \midrule
  \multirow{1}{*}{SSAF-CN~\cite{li2022dse}}  & \multirow{2}{*}{text effects} & \multirow{1}{*}{97,200}  & \multirow{1}{*}{100 text effects rendered on 972 Chinese characters.} \\   
  \multirow{1}{*}{SSAF-EN~\cite{li2022dse}}  &  & \multirow{1}{*}{2,600}  & \multirow{1}{*}{100 text effects rendered on 26 English letters.} \\    
  
  \midrule
  \multirow{1}{*}{Imgur5K~\cite{krishnan2023textstylebrush}}  & \multirow{1}{*}{handwriting} & \multirow{1}{*}{135,375}  & \multirow{1}{*}{135,375 handwritten English words from 5,305 images} \\   

  \midrule
  \multirow{2}{*}{TextLogo3K~\cite{wang2022aesthetic}}  & \multirow{2}{*}{text logo} & \multirow{2}{*}{3,470}  & \multirow{1}{*}{text logo images extracted from} \\  
  & & & \multirow{1}{*}{poster/covers of movies, TV series and comics.}\\  

  \midrule
  \multirow{2}{*}{MARIO-10M~\cite{chen2023textdiffuser}}  & \multirow{2}{*}{visual text} & \multirow{2}{*}{10,061,720}  & \multirow{1}{*}{9,194,613, 343,423 and 523,684 text images from } \\  
  & & & \multirow{1}{*}{natural images, posters, and book covers, respectively.}\\  
  
  \midrule
  \multirow{1}{*}{LAION-Glyph~\cite{yang2024glyphcontrol}}  & \multirow{1}{*}{visual text} & \multirow{1}{*}{$\sim$10,000,000}  & \multirow{1}{*}{images with rich visual text content} \\  

  \midrule
  \multirow{3}{*}{AnyWord-3M~\cite{tuo2023anytext}}  & \multirow{3}{*}{visual text} & \multirow{3}{*}{3,034,486}  & \multirow{1}{*}{text images from several datasets} \\  
  & & & \multirow{1}{*}{1.6 million in Chinese, 1.39 million in English,}\\ 
  & & & \multirow{1}{*}{and 10k images in other languages.}\\ 
\bottomrule
\end{tabular}}
\label{tab:dataset}
\end{table*}

\section{Dataset and Evaluation}
\label{sec6}

\subsection{Datasets}

\begin{figure}[htbp]
    \centering
    \includegraphics[width=1\linewidth]{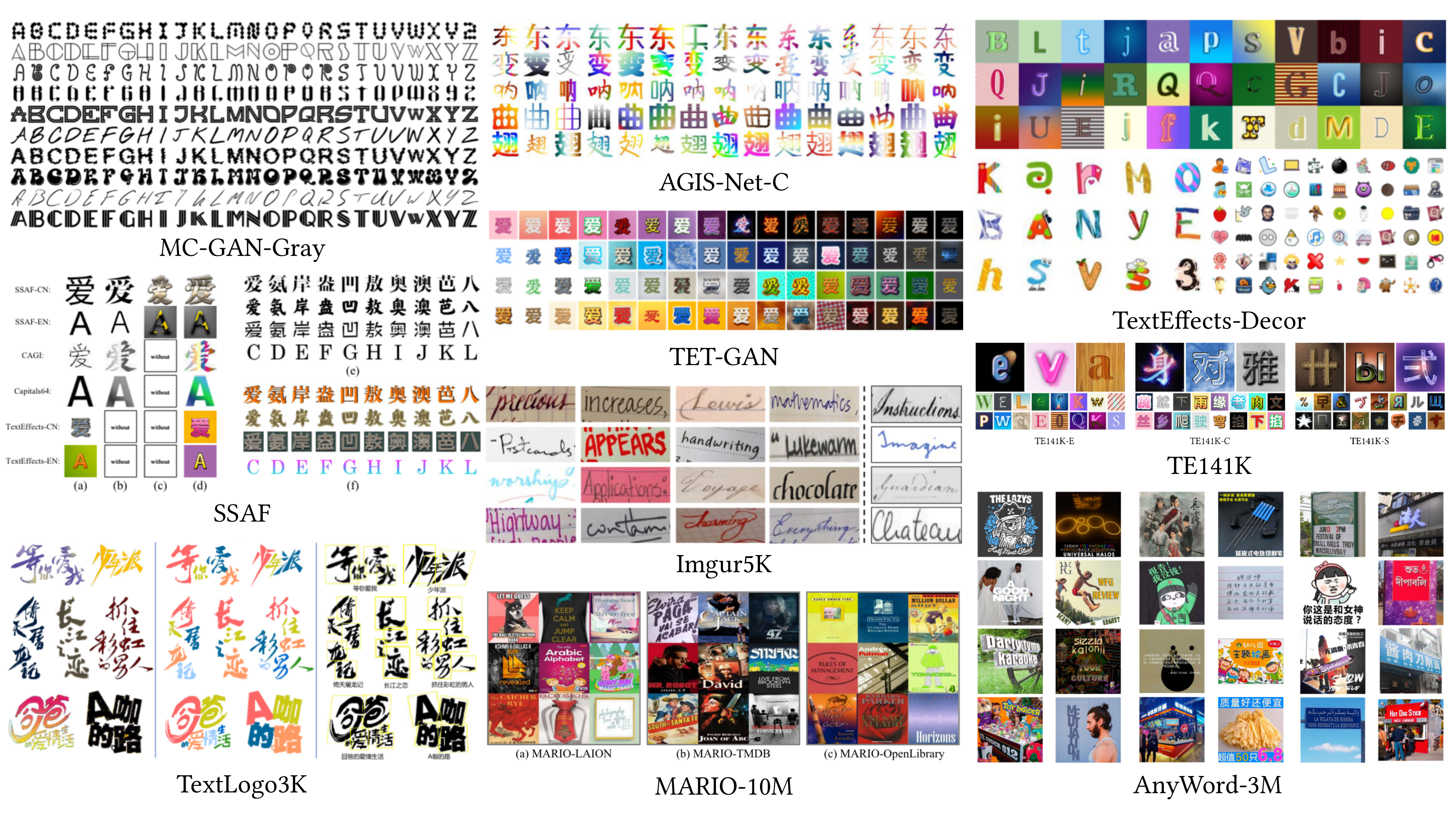}
    \caption{Samples of the datasets for artistic text rendering and design.}
    \label{fig:datasets}
\end{figure}

As summarized in Table 2, there are multiple available datasets for artistic text rendering and design. As shown in Figure~\ref{fig:datasets}, these datasets mainly differ in several major aspects: 1) whether font styles are taken into consideration, 2) whether the reference text style contains special elements, 3) the kinds of character types that are involved, and 4) the usage scenarios in the real world. 

For the first aspect, the first four datasets, namely MC-GAN-Gray~\cite{azadi2018multi}, MC-GAN-Color~\cite{azadi2018multi}, AGIS-Net-C~\cite{gao2019artistic}, and AGIS-Net-P~\cite{gao2019artistic}, consider fonts along with text effects. For the second aspect, only TextEffects-Decor~\cite{wang2019typography} collects effects with decorative elements. For the third aspect, TET-GAN~\cite{yang2019tet}, TE141K-C~\cite{yang2020te141k}, TE141K-S~\cite{yang2020te141k}, and AnyWord-3M~\cite{tuo2023anytext} contain characters of multiple languages, while others only collect either English letters or Chinese characters. For the last aspect, Imgur5K~\cite{krishnan2023textstylebrush}, TextLogo3K~\cite{wang2022aesthetic}, MARIO-10M~\cite{chen2023textdiffuser}, LAION-Glyph~\cite{yang2024glyphcontrol}, and AnyWord-3M presenting text images from various life circumstances, spanning aspects of advertising billboards, movie posters, book covers, handwritten words, internet memes, \etc.

In the following, we briefly introduce the details of the representative datasets for artistic text rendering and design.

\begin{enumerate}
    \item MC-GAN-Gray~\cite{azadi2018multi} is a joint font and text effects dataset that includes 10K gray-scale Latin fonts each with 26 capital English letters, a total of 260,000 images. These images were processed by bounding boxes, and each glyph was standardized and placed at the center of a 64 × 64 image. Excluding font ornamentation, these fonts provide extensive information about inter-letter correlations in font styles, encoding only the glyph outlines. Based on the former, Azadi \etal~apply random color gradients and outlining onto the gray-scale glyphs, resulting in a dataset of 20K color fonts and a total of 520,000 images, namely MC-GAN-Color~\cite{azadi2018multi}.
    \item AGIS-Net-C~\cite{gao2019artistic} dataset is a joint font and text effects dataset that includes 246 normal Chinese font styles and 10 different kinds of gradient colors and stripe textures, with 639 representative Chinese characters, a total of 1,571,940 images. For few-shot tasks, Gao \etal~also introduce the AGIS-Net-P~\cite{gao2019artistic} dataset, which consists of 35 selected professional-designed fonts with 7,326 Chinese characters for each style, a total of 256,410 images.
    \item TET-GAN~\cite{yang2019tet} dataset is a text effects dataset that includes 64 text effects each with 775 Chinese characters, 52 English letters, and 10 Arabic numerals, a total of 53,568 images. Each text effects image, sized at 320 × 320, is accompanied by its corresponding plain text image. 
    \item TextEffects-Decor~\cite{wang2019typography} is a text effects dataset that includes 60 text effects on 52 English letters of 19 fonts, a total of 59,280 images. Each text effect has a size of 320 × 320 and is provided with its corresponding raw text, and decorative elements were collected from \url{www.shareicon.net}.
    \item TE141K~\cite{yang2020te141k} is a text effects dataset that includes 152 text effects rendered on glyphs (English letters, Chinese characters, and Arabic numerals). Each text effect has a resolution of 320 × 320 and is provided with its corresponding glyph image. Based on glyph types, TE141K was divided into three subsets: TE141K-E contains 67 styles and 988 glyphs, totally 66,196 image pairs of English alphabets; TE141K-C contains 65 styles and 837 glyphs, totally 54,405 image pairs of Chinese characters; TE141K-S contains 20 styles and 1,024 glyphs, totally 20,480 image pairs of special symbols and letters from common languages other than Chinese and English.
    \item SSAF~\cite{li2022dse} is a text effects dataset that includes 200 text effects rendered on glyphs (Chinese characters and English letters).  Each text effect has a resolution of 320 × 320 and is provided with its corresponding glyph image. Based on glyph types, SSAF was divided into two subsets: SSAF-CN contains 100 Chinese artistic fonts, each with 972 Chinese characters; SSAF-EN contains 100 English artistic fonts, each with 26 uppercase English letters.
    \item Imgur5K~\cite{krishnan2023textstylebrush} is a handwriting image dataset that includes 135,375 handwritten English words from 5K images originally hosted publicly on \url{Imgur.com}. Each image was assigned to no more than five annotators, and spurious data was eliminated by the utility of the annotation averages of word bounding boxes and the highest agreement on labeled content strings.
    \item TextLogo3K~\cite{wang2022aesthetic} is a text logo dataset that includes 3,470 text logo images of posters and covers of movies, TV series, and comics which were selected from Tencent Video, one of the leading online video platforms in China. Each character has been annotated by bounding box, pixel-level mask, category, and the angle of rotation and affine transformation  (if existing). 
    \item MARIO-10M~\cite{chen2023textdiffuser} is a visual text dataset that includes 10,061,720 image-text pairs of natural images, posters, and book covers, each with comprehensive OCR annotations. Based on the difference in data sources, MARIO-10K was divided into three subsets:  MARIO-LAION contains 9,194,613 high-quality text images with corresponding captions, covering a broad spectrum of advertisements, notes, posters, covers, memes, logos, \etc;  MARIO-TMDB contains 343,423 English posters from The Movie Database (TMDB), a community-built dataset for movies and TV shows with high-quality posters; MARIO-OpenLibrary contains 523,684 original-size covers from Open Library, an open and editable library catalog that generates a web page for every published book.
    \item LAION-Glyph~\cite{yang2024glyphcontrol} is a visual text dataset that includes about 10M images with rich visual text content which were selected by using a modern OCR system. The amount of characters in the images primarily ranges from 10 to 50, with most samples containing fewer than 150 characters. Based on the considerations for the convenience of training and evaluation, the LAION-Glyph dataset was divided into three scales: LAION-Glyph-100K, LAION-Glyph-1M, and LAION-Glyph-10M.
    \item AnyWord-3M~\cite{tuo2023anytext} is a visual text dataset that includes 3,034,486 multi-language scene text images, covering street views, book covers, advertisements, posters, movie frames, \etc. AnyWord-3M contains approximately 1.6M images in Chinese, 1.39M images in English, and 10k images in other languages, spanning Japanese, Korean, Arabic, Bengali, and Hindi. 
\end{enumerate}

\subsection{Performance Evaluation}

For text style transfer tasks with paired output and ground truth, instance-level similarities such as L1 loss, MSE loss, peak signal-to-noise ratio (PSNR), structural similarity index (SSIM), and perceptual loss~\cite{johnson2016perceptual} can be used to calculate differences between the output image and the ground truth image. Inception score (IS)~\cite{radford2015unsupervised} and Fréchet inception distance (FID)~\cite{salimans2016improved} are introduced to measure the distribution distance between the generated images and ground truth dataset. However, the aforementioned evaluation metrics are mostly designed for natural images, thus limiting the evaluation ability of artistic text.

To tailor the evaluation metrics to the realm of artistic text, content- and style-related metrics are introduced to this task. Style loss~\cite{johnson2016perceptual} and CLIP score~\cite{radford2021learning} are widely used in semantic character generation and visual text generation for further assessing the expression of semantic information. To evaluate the text legibility and correctness, OCR accuracy, Sentence Accuracy, and Text Detection Accuracy are mainly used in scene text editing tasks~\cite{tuo2023anytext,zhang2024brush,yang2020swaptext,qu2023exploring,wu2019editing}. Yan \etal~\cite{yan2020multitask} provide a novel assessment system specially focused on artistic text stylization. A multi-task network is proposed to extract features of artistic text images and is trained on selected data from TE141K~\cite{yang2020te141k} with user labels, thus imitating visual evaluation from humans. This model stands as a robust instrument to assist the quality assessment process.

Qualitative evaluation methods such as user study are effective ways to analyze the quality and aesthetic appeal of the generated result by human aesthetic judgment. However, the result could be influenced by the preference of individual participants, which gives it a certain degree of subjectivity and makes it difficult to reproduce.

\section{Future Challenges}
\label{sec7}

Although the rapid advancement of artificial intelligence and deep learning has made great progress in artistic text generation and design, and the recent methods can generate satisfactory results, there are still several challenges and open issues. This section discusses some key challenges in artistic text generation.

One of the challenges in the automatic generation of artistic text is that current methods stylize text best based on concrete visual concepts. It is still difficult to use abstract concepts to influence the stylization process. This limitation hinders the ability to fully leverage the creative potential of abstract concepts and ideas in the artistic rendering of text. One possible solution is to leverage large language models (LLMs) to rephrase the abstract concepts into more descriptive ones. 

Another challenge is that current methods are mostly based on diffusion models. However, the sampling process of diffusion models is well-known to be inherently slow, leading to a slow generation of artistic text. A possible solution is to explore faster approximation methods or efficient sampling techniques. Additionally, model distillation could further reduce generation time.

While significant progress has been made in static artistic text generation with various emerging methods, dynamic artistic text generation remains under-explored. This is partly due to the lack of sufficient video data and the increased complexity of generating video compared to images. One great demand is to develop large, high-quality datasets specifically for dynamic artistic text. Additionally, employing advanced video generation techniques, such as temporal consistency models and leveraging transfer learning from static to dynamic scenarios, could help address the challenges in this area.

Another challenge in artistic text generation is achieving fine-grained control. While diffusion-based methods can generate diverse artistic text, text guidance offers coarse-grained control over the output. It is difficult to achieve fine-grained control, such as changing specific regions or styles of the text, adjusting individual character shapes, adapting the text to different font sizes, or flexibly modifying the artistic degree to balance artistry and legibility. A potential solution is to integrate attention mechanisms and region-specific conditioning into the models. Additionally, developing interactive tools that allow users to manually adjust and refine specific aspects of the generated artistic text can also enhance fine-grained control and customization.

\section{Conclusion}
\label{sec8}

This paper provides a comprehensive survey of artistic text rendering and design. 
We first explored two main categories: artistic text stylization and semantic typography, along with the incorporation of motion for dynamic artistic text generation.
Second, several applications in the context of artistic text generation are detailed. 
Third, we studied the datasets and evaluation of the artistic text.
Finally, several challenges and future research directions were discussed. 
The need for creative, efficient, versatile, and controllable generative models was emphasized.
We hope this paper can serve as a foundation for further advancements and inspire the exploration of new avenues in this exciting field.

\bibliographystyle{splncs04}
\bibliography{main}

\end{document}